\documentclass[oneside,twocolumn, 10pt,DIV21]{scrartcl}

\usepackage{listings}
\usepackage{color}
\usepackage{evolvingai}
\usepackage{natbib}
\usepackage{graphicx}
\usepackage{lettrine}
\usepackage{url}
\date{}
\usepackage{float}
\newfloat{source}{tbhp}{lst}
\floatname{source}{Source Code}

\definecolor{mygreen}{rgb}{0,0.6,0}
\definecolor{mygray}{rgb}{0.5,0.5,0.5}
\definecolor{mymauve}{rgb}{0.58,0,0.82}

\usepackage{lipsum}

\usepackage{courier}
  \usepackage{caption}
\DeclareCaptionFont{white}{\color{white}}
\DeclareCaptionFormat{listing}{\colorbox[cmyk]{0.43, 0.35, 0.35,0.01}{\parbox{\textwidth}{\hspace{15pt}#1#2#3}}}
\title{\center{\fontsize{19}{25}\selectfont \textsf{\textbf{Limbo: A Fast and Flexible Library for Bayesian Optimization}}
  \textbf{\normalsize{\textsf{
      Antoine Cully$\mathsf{^{1}}$,
      Konstantinos Chatzilygeroudis$\mathsf{^{2,3,4}}$,
      Federico Allocati$\mathsf{^{2,3,4}}$
      and Jean-Baptiste Mouret$\mathsf{^{2,3,4}}$}}}}
      \flushleft{
         \small{
         $\mathsf{^1}$ Personal Robotics Lab, Imperial College London, London, UK\\
         $\mathsf{^2}$ Inria Nancy Grand - Est, Villers-l\`es-Nancy, France\\
         $\mathsf{^3}$ CNRS, Loria, UMR 7503,  Vand\oe{}uvre-l\`es-Nancy, France\\
        $\mathsf{^4}$ Universit\'e de Lorraine, Loria, UMR 7503, Vand\oe{}uvre-lès-Nancy, France\\
}
~\\
\noindent{\normalsize \textsf{Correspondence to: jean-baptiste.mouret@inria.fr}}\\
\noindent{\normalsize \textsf{Preprint -- \today}}
\vspace*{-1.0cm}
}}

\begin{document}





\maketitle

\begin{abstract}
\bfseries\sffamily
\noindent{}Limbo is an open-source C++11 library for Bayesian optimization which is designed to be both highly flexible and very fast. It can be used to optimize functions for which the gradient is unknown, evaluations are expensive, and runtime cost matters (e.g., on embedded systems or robots). Benchmarks on standard functions show that Limbo is about 2 times faster than BayesOpt (another C++ library) for a similar accuracy.
\end{abstract}

\section*{Introduction}
\lettrine{N}{on-linear} optimization problems are pervasive in machine learning.
Bayesian Optimization (BO) is designed for the most challenging ones: when the gradient is unknown, evaluating a solution is costly, and evaluations are noisy. This is, for instance, the case when we want to find optimal parameters for a machine learning algorithm \citep{snoek2012practical}, because testing a set of parameters can take hours, and because of the stochastic nature of many machine learning algorithms. Besides parameter tuning, Bayesian optimization recently attracted a lot of interest for direct policy search in robot learning \citep{lizotte2007automatic,wilson2014using, calandra2016bayesian} and online adaptation; for example, it was recently used to allow a legged robot to learn a new gait after a mechanical damage in about 10-15 trials (2 minutes) \citep{cully2015robots}.

At its core, Bayesian optimization builds a probabilistic model of the function to be optimized (the reward/performance/cost function) using the samples that have already been evaluated \citep{shahriari2016taking}; usually, this model is a Gaussian process \citep{williams2006gaussian}. To select the next sample to be evaluated, Bayesian optimization optimizes an \emph{acquisition function} which leverages the model to predict the most promising areas of the search space. Typically, this acquisition function is high in areas not yet explored by the algorithm (i.e., with a high uncertainty) and in those where the model predicts high-performing solutions. As a result, Bayesian optimization handles the exploration / exploitation trade-off by selecting samples that combine a good predicted value and a high uncertainty.

In spite of its strong mathematical foundations \citep{mockus2012bayesian}, Bayesian optimization is more a template than a fully-specified algorithm. For any Bayesian optimization algorithm, the following components need to be chosen: (1) an initialization function (e.g., random sampling), (2) a model (e.g., a Gaussian process, which itself needs a kernel function and a mean function), (3) an acquisition function (e.g., Upper Confidence Bound, Expected Improvement, see \citealt{shahriari2016taking}), (4) a global, non-linear optimizer for the acquisition function (e.g., CMA-ES, \citealt{hansen2001completely}, or DIRECT, \citealt{jones1993lipschitzian}) (5) a non-linear optimizer to learn the hyper-parameters of the models (if the user chooses to learn them). Specific applications often require a specific choice of components and most research articles focus on the introduction of a single component (e.g., a novel acquisition function or a novel kernel for Gaussian processes).

This almost infinite number of variants of Bayesian optimization calls for flexible libraries in which components can easily be substituted with alternative ones (user-defined or predefined). In many applications, the run-time cost is negligible compared to the evaluation of a potential solution, but this is not the case in online adaptation for robots (e.g., \citealt{cully2015robots}), in which the algorithm has to run on small embedded platforms (e.g., a cell phone), or in interactive applications \citep{brochu2010tutorial}, in which the algorithm needs to quickly react to the inputs. To our knowledge, no open-source library combines a high-performance implementation of Bayesian optimization with the high flexibility that is needed for developing and deploying novel variants.
\begin{figure*}[!ht]
  \centering
  \includegraphics[width=0.95\textwidth]{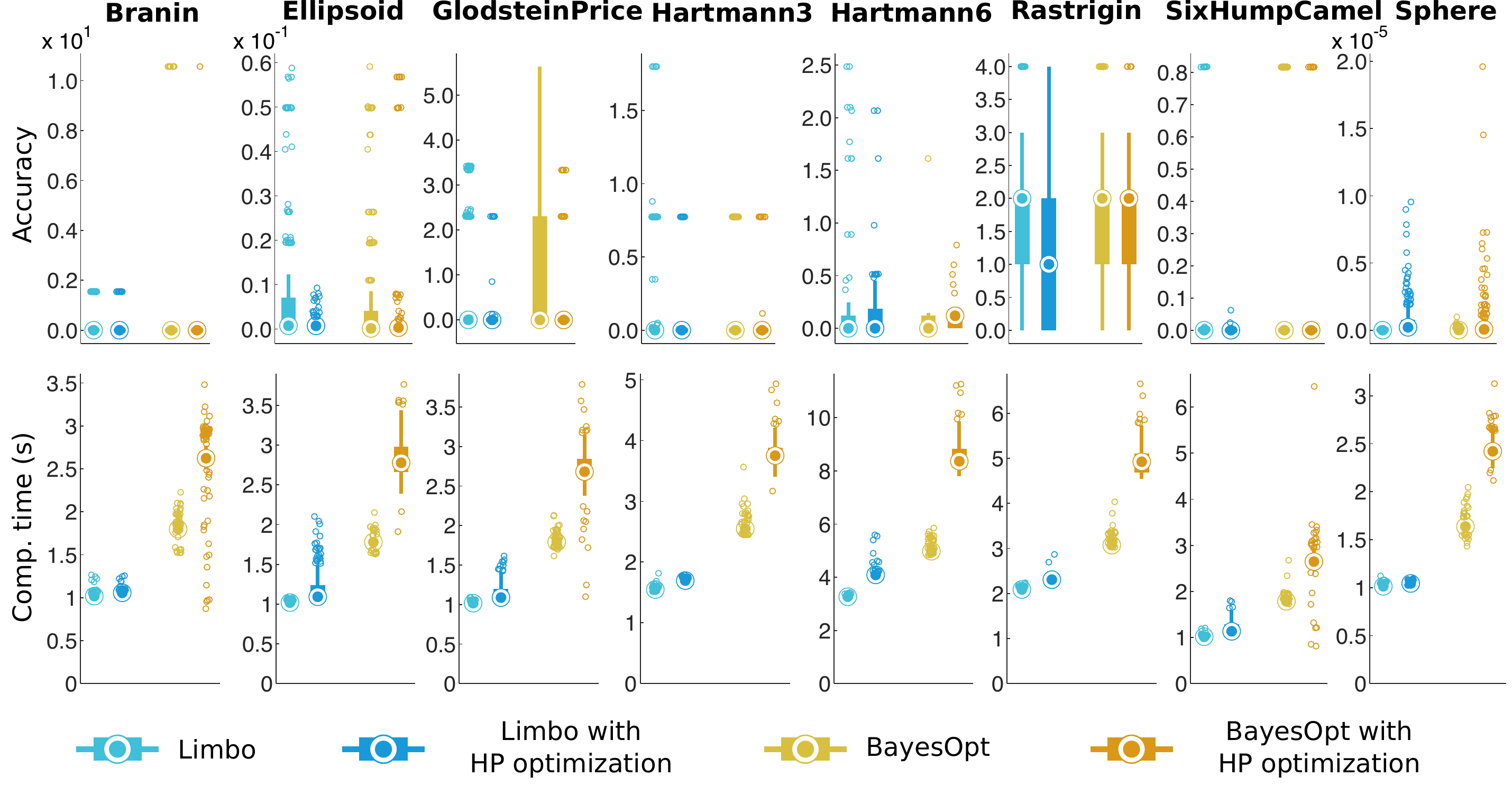}
  \caption[]{\label{fig:bench_res}Comparison of the accuracy (difference with the optimal value) and the
    wall clock time for Limbo and BayesOpt \citep{martinezcantin14a} -- a state-of-the art C++ library for Bayesian optimization -- on common test
    functions (see \url{http://www.sfu.ca/~ssurjano/optimization.html}). Two configurations are
    tested: with and without optimization of the hyper-parameters
    of the Gaussian Process. Each experiment has been replicated 250
    times. The median of the data is pictured with a thick dot, while
    the box represents the first and third quartiles. The most extreme
    data points are delimited by the whiskers and the outliers are
    individually depicted as smaller circles. Limbo is configured to
    reproduce the default parameters of BayesOpt.}
\end{figure*}

\section*{The Limbo library}
Limbo (LIbrary for Model-based Bayesian Optimization) is an open-source (GPL-compatible CeCiLL license) C++11 library which provides a modern implementation of Bayesian optimization that is both flexible and high-performing. It does not depend on any proprietary software (the main dependencies are Boost and Eigen3). The code is standard-compliant but it is currently mostly developed for GNU/Linux and Mac OS X with both the GCC and Clang compilers.
The library is distributed via a GitHub repository\footnote{\url{http://github.com/resibots/limbo}}, in which bugs and further developments are handled by the community of developers and users. An extensive documentation\footnote{\url{http://www.resibots.eu/limbo}} is available and contains guides, examples, and tutorials. New contributors can rely on a full API reference, while their developments are checked via a continuous integration platform (automatic unit-testing routines). Limbo was instrumental in several of our robotics projects (e.g., \citealt{cully2015robots}) but it has successfully been used internally for other fields.



The implementation of Limbo follows
a template-based, \emph{policy-based design} \citep{alexandrescu2001modern}, which allows it to be highly flexible without paying the cost induced by classic object-oriented designs \citep{driesen1996direct} (cost of virtual functions). In practice, changing one of the components of the algorithms in Limbo (e.g., changing the acquisition function) usually requires changing only a template definition in the source code. This design make it possible for users to rapidly experiment and test new ideas while being as fast as specialized code.


According to the benchmarks we performed (Figure \ref{fig:bench_res}), Limbo
finds solutions with the same level of quality as BayesOpt \citep{martinezcantin14a},
within a significantly lower amount of time: for the same accuracy (less than $2.10^{-3}$ between the optimized solutions found by Limbo and BayesOpt),
Limbo is between $1.47$ and $1.76$ times faster (median
values) than BayesOpt when the hyper-parameters are not optimized, and
between $2.05$ and $2.54$ times faster when they are.

\section*{Using Limbo}
The policy-based design of Limbo allows users to define or adapt
variants of Bayesian Optimization with very little change in the
code. The definition of the optimized function is achieved by
creating a functor (an arbitrary object with an \emph{operator()}
function) that takes as input a vector and outputs the resulting
vector (Limbo can support multi-objective optimization); this object also defines the input and output dimensions of
the problem (\textsf{dim\_in}, \textsf{dim\_out}). For example, to maximize the function $\textsf{my\_fun}(\textbf{x}) = -\sum_{i=1}^{2} x_i^2 \sin(2 x_i)$:
~\\

\begin{lstlisting}[language=C++]
struct my_fun {
    static constexpr size_t dim_in = 2;
    static constexpr size_t dim_out = 1;
    Eigen::VectorXd operator()(const Eigen::VectorXd& x) const {
        double res = -(x.array().square() * (x * 2).sin()).sum();
        return limbo::tools::make_vector(res);
    }
};
\end{lstlisting}
Optimizing \textsf{my\_fun} with the default parameters only requires instantiating a BOptimizer object and call the ``optimize'' method:
\begin{lstlisting}
limbo::bayes_opt::BOptimizer<Params> opt;
opt.optimize(my_fun());
\end{lstlisting}
where \textsf{Params} is a structure that defines all the parameters in a static way, for instance:
\begin{lstlisting}[language=C++]
struct Params {
  // default parameters for the acquisition function 'gpucb'
  struct acqui_gpucb : public limbo::defaults::acqui_gpucb { };
  // custom parameters for the optimizer
  struct bayes_opt_boptimizer : public limbo::defaults::bayes_opt_boptimizer {
    BO_PARAM(double, noise, 0.001);
  };
  // ...
}
\end{lstlisting}

While default functors are provided, most of the components of \emph{BOptimizer} can be replaced to allow researchers to investigate new variants. For example, changing the kernel function from the Squared Exponential kernel (the default) to another type of kernel (here the Matern-5/2) and using the UCB acquisition function is achieved as follows:
\begin{lstlisting}[language=C++]
// define the templates
using Kernel_t = limbo::kernel::MaternFiveHalves<Params>;
using Mean_t = limbo::mean::Data<Params>;
using GP_t = limbo::model::GP<Params, Kernel_t, Mean_t>;
using Acqui_t = limbo::acqui::UCB<Params, GP_t>;
// instantiate a custom optimizer
limbo::bayes_opt::BOptimizer<Params, limbo::modelfun<GP_t>, limbo::acquifun<Acqui_t>> opt;
// run it
opt.optimize(my_fun());
\end{lstlisting}





In addition to the many kernel, mean, and acquisition functions that
are implemented, Limbo provides several tools for the internal
optimization of the acquisition function and the hyper-parameters. For
example, a wrapper around the NLOpt library (which provides many
local, global, gradient-based, and gradient-free algorithms) allows Limbo
to be used with a large variety of internal optimization
algorithms. Moreover, several ``restarts'' of these internal
optimization processes can be performed in parallel to avoid local
optima with a minimal computational cost, and several internal
optimizations can be chained in order to take advantage of the global
aspects of some algorithms and the local properties of others.




\section*{Acknowledgements}
{This project is funded by the European Research Council (ERC) under the European Union's Horizon 2020 research and innovation programme (Project: ResiBots, grant agreement No 637972)}.

%
%
%
%
%
%
%
%
%
\bibliographystyle{plainnat}
\sffamily
\bibliography{biblio}

\begin{thebibliography}{14}
\providecommand{\natexlab}[1]{#1}
\providecommand{\url}[1]{\texttt{#1}}
\expandafter\ifx\csname urlstyle\endcsname\relax
  \providecommand{\doi}[1]{doi: #1}\else
  \providecommand{\doi}{doi: \begingroup \urlstyle{rm}\Url}\fi

\bibitem[Alexandrescu(2001)]{alexandrescu2001modern}
Andrei Alexandrescu.
\newblock \emph{Modern {C++} design: generic programming and design patterns
  applied}.
\newblock Addison-Wesley, 2001.

\bibitem[Brochu et~al.(2010)Brochu, Cora, and De~Freitas]{brochu2010tutorial}
Eric Brochu, Vlad~M Cora, and Nando De~Freitas.
\newblock A tutorial on {Bayesian} optimization of expensive cost functions,
  with application to active user modeling and hierarchical reinforcement
  learning.
\newblock \emph{arXiv preprint arXiv:1012.2599}, 2010.

\bibitem[Calandra et~al.(2016)Calandra, Seyfarth, Peters, and
  Deisenroth]{calandra2016bayesian}
Roberto Calandra, Andr{\'e} Seyfarth, Jan Peters, and Marc~Peter Deisenroth.
\newblock {Bayesian} optimization for learning gaits under uncertainty.
\newblock \emph{Annals of Mathematics and Artificial Intelligence}, 76\penalty0
  (1-2):\penalty0 5--23, 2016.

\bibitem[Cully et~al.(2015)Cully, Clune, Tarapore, and Mouret]{cully2015robots}
Antoine Cully, Jeff Clune, Danesh Tarapore, and Jean-Baptiste Mouret.
\newblock Robots that can adapt like animals.
\newblock \emph{Nature}, 521\penalty0 (7553):\penalty0 503--507, 2015.

\bibitem[Driesen and H{\"o}lzle(1996)]{driesen1996direct}
Karel Driesen and Urs H{\"o}lzle.
\newblock The direct cost of virtual function calls in {C++}.
\newblock In \emph{ACM Sigplan Notices}, volume~31, pages 306--323. ACM, 1996.

\bibitem[Hansen and Ostermeier(2001)]{hansen2001completely}
Nikolaus Hansen and Andreas Ostermeier.
\newblock Completely derandomized self-adaptation in evolution strategies.
\newblock \emph{Evolutionary computation}, 9\penalty0 (2):\penalty0 159--195,
  2001.

\bibitem[Jones et~al.(1993)Jones, Perttunen, and
  Stuckman]{jones1993lipschitzian}
Donald~R Jones, Cary~D Perttunen, and Bruce~E Stuckman.
\newblock {Lipschitzian} optimization without the {Lipschitz} constant.
\newblock \emph{Journal of Optimization Theory and Applications}, 79\penalty0
  (1):\penalty0 157--181, 1993.

\bibitem[Lizotte et~al.(2007)Lizotte, Wang, Bowling, and
  Schuurmans]{lizotte2007automatic}
Daniel~J Lizotte, Tao Wang, Michael~H Bowling, and Dale Schuurmans.
\newblock Automatic gait optimization with {Gaussian} process regression.
\newblock In \emph{IJCAI}, volume~7, pages 944--949, 2007.

\bibitem[Martinez-Cantin(2014)]{martinezcantin14a}
Ruben Martinez-Cantin.
\newblock {BayesOpt:} a {Bayesian} optimization library for nonlinear
  optimization, experimental design and bandits.
\newblock \emph{Journal of Machine Learning Research}, 15:\penalty0 3915--3919,
  2014.

\bibitem[Mockus(2012)]{mockus2012bayesian}
Jonas Mockus.
\newblock \emph{{Bayesian} approach to global optimization: theory and
  applications}, volume~37.
\newblock Springer Science \& Business Media, 2012.

\bibitem[Shahriari et~al.(2016)Shahriari, Swersky, Wang, Adams, and
  de~Freitas]{shahriari2016taking}
Bobak Shahriari, Kevin Swersky, Ziyu Wang, Ryan~P Adams, and Nando de~Freitas.
\newblock Taking the human out of the loop: A review of {Bayesian}
  optimization.
\newblock \emph{Proceedings of the IEEE}, 104\penalty0 (1):\penalty0 148--175,
  2016.

\bibitem[Snoek et~al.(2012)Snoek, Larochelle, and Adams]{snoek2012practical}
Jasper Snoek, Hugo Larochelle, and Ryan~P Adams.
\newblock Practical {Bayesian} optimization of machine learning algorithms.
\newblock In \emph{Advances in neural information processing systems}, pages
  2951--2959, 2012.

\bibitem[Williams and Rasmussen(2006)]{williams2006gaussian}
Christopher~KI Williams and Carl~Edward Rasmussen.
\newblock Gaussian processes for machine learning.
\newblock \emph{the MIT Press}, 2\penalty0 (3):\penalty0 4, 2006.

\bibitem[Wilson et~al.(2014)Wilson, Fern, and Tadepalli]{wilson2014using}
Aaron Wilson, Alan Fern, and Prasad Tadepalli.
\newblock Using trajectory data to improve {Bayesian} optimization for
  reinforcement learning.
\newblock \emph{Journal of Machine Learning Research}, 15\penalty0
  (1):\penalty0 253--282, 2014.

\end{thebibliography}

\end{document}